%% file: ijcai24.tex
\documentclass{article}
\usepackage{ijcai24}

\usepackage{times}
\usepackage{soul}
\usepackage{url}
\usepackage[hidelinks]{hyperref}
\usepackage[utf8]{inputenc}
\usepackage[small]{caption}
\usepackage{graphicx}
\usepackage{amsmath}
\usepackage{amssymb}
\usepackage{amsthm}
\usepackage{booktabs}
\usepackage{algorithm}
\usepackage{algorithmic}
\usepackage{multirow}
\usepackage[switch]{lineno}
\usepackage{todonotes}

\title{PathE: Leveraging Entity-Agnostic Paths for \\Parameter-Efficient Knowledge Graph Embeddings}

\author{
Ioannis Reklos$^1$
\and
Jacopo de Berardinis$^2$\and
Elena Simperl$^{1}$\And
Albert Meroño-Peñuela$^1$\\
\affiliations
$^1$King's College London\\
$^2$University of Liverpool\\
\emails
\{ioannis.reklos, elena.simperl,albert.merono\}@kcl.ac.uk,
jacodb@liverpool.ac.uk
}

\begin{document}

\newcommand{\W}[1]{{\textcolor{red}{#1}}}
\newcommand{\ioannis}[1]{{\textcolor{orange}{\textbf{I}: #1}}}
\newcommand{\albert}[1]{{\textcolor{brown}{\textbf{A}: #1}}}
\newcommand{\elena}[1]{{\textcolor{magenta}{\textbf{E}: #1}}}
\newcommand{\jaco}[1]{{\textcolor{blue}{\textbf{J}: #1}}}

\maketitle

\begin{abstract} 
Knowledge Graphs (KGs) store human knowledge in the form of entities (nodes) and relations, and are used extensively in various applications.
KG embeddings are an effective approach to addressing tasks like knowledge discovery, link prediction, and reasoning.
This is often done by allocating and learning embedding tables for all or a subset of the entities.
As this scales linearly with the number of entities, learning embedding models in real-world KGs with millions of nodes can be computationally intractable.
To address this scalability problem, our model, \textbf{PathE}, only allocates embedding tables for relations (which are typically orders of magnitude fewer than the entities) and requires less than 25\% of the parameters of previous parameter efficient methods.
Rather than storing entity embeddings, we learn to compute them by leveraging multiple entity-relation paths to contextualise individual entities within triples.
Evaluated on four benchmarks, PathE achieves state-of-the-art performance in relation prediction, and remains competitive in link prediction on path-rich KGs while training on consumer-grade hardware.
We perform ablation experiments to test our design choices and analyse the sensitivity of the model to key hyper-parameters.
PathE is efficient and cost-effective for relationally diverse and well-connected KGs commonly found in real-world applications.

\end{abstract}

\section{Introduction}\label{sec:introduction}

Knowledge Graphs (KGs) such as Wikidata and Freebase serve as a structured embodiment of human knowledge in machine readable format.
They consist of a large number of \texttt{(subject, relation, predicate)} triples, where subject and predicate (alias \textit{head} and \textit{tail}) are nodes in the KG and the relation is the edge connecting them.
Each triple denotes an atomic fact, such as \texttt{(London, capital\_Of, England)}.
KGs are ubiquitous and are used in question answering, information retrieval~\cite{survey_kg_applications}, recommendation systems~\cite{DKG_rec_systems} and autonomous agents~\cite{KG_autonomous_robots}, and they can augment Large Language Models (LLMs) with facts and common sense knowledge~\cite{LLM_KG} from authoritative sources.
However, KGs are usually incomplete, which means that information (nodes and edges) is missing from the graph, and new triples are added.
An effective method for KG completion is learning Knowledge Graph Embeddings (KGE), either by storing the learned representations in embedding tables~\cite{Rotate,TransE} or by utilising more complex Graph Neural Network (GNN) architectures to leverage their inherent structure~\cite{compgcn,REDGNN}.
Embeddings provide good performance in KG completion tasks~\cite{BoxE,ConvE} but suffer from significant drawbacks: their computation may become intractable on large web-scale KGs ($10^{6-9}$ nodes); and they struggle to embed unseen nodes, called \textit{inductive embedding}, without being retrained from scratch.
Furthermore, state of the art methods based on GNNs, such as CompGCN \cite{compgcn}, require storing the whole adjacency matrix which limits their applicability to larger KGs \cite{REDGNN}. Specialised architectures \cite{REDGNN,NBFnets} have attempted to address these issues by producing entangled representations of node pairs, thereby removing the need for storing the adjacency matrix at the expense of producing node-level representations.

Recent work has also explored reducing the memory requirements of KGE by focusing on the relations between entities.
\citeauthor{nodepiece} achieve scalability by allocating embedding tables only for a subset of entities (alias \textit{anchors}) and encode the others based on their distance from the anchors.

\citeauthor{EARL} leverage a fixed vocabulary of embedded nodes (alias \textit{reserved entities}) and relational context similarity, in conjunction with a GNN model, to improve performance and retain inductive capabilities.
Nonetheless, these efforts still require storing embedding tables for reserved nodes.

To overcome these limitations, we introduce \textbf{PathE}, a parameter-efficient KGE method that departs from traditional approaches by storing only relation representations and dynamically computing entity embeddings.
PathE leverages path information to contextualise nodes and their connectivity patterns, generating structure-aware entity representations without the computational overhead of message passing in GNNs, nor utilising any stored node representations.

Specifically, paths are drawn from unique random walks starting or ending from/at each entity. 

Entities are encoded via their relational context, which is defined as the number and type of the relations they appear with (either as head or tail, for outgoing and incoming contexts respectively).
A node projector learns to project entity-specific relational contexts by forwarding this information through a series of fully connected layers; which yields an entity representation that has the same dimension of the relation embeddings.
Entity-relation paths are then constructed by combining node projections and embeddings, respectively.
Given a triple $(h, r, t)$, multiple incoming and outgoing paths for each entity $(h, t)$ are processed by a sequence model, and an aggregation strategy is applied across all the entity representations in each path.
This yields separate embedding vectors for head, tail, and relation -- which are trained using a learning objective for link prediction or relation prediction.
Overall, this provides a more scalable and inductive solution, as it allows the model to embed new/unseen entities without retraining.

Through extensive empirical evaluation on various KG benchmarks, we demonstrate PathE's effectiveness as a novel parameter-efficient KGE method.
It achieves state of the art performance in \textit{relation prediction} and remains competitive in \textit{link prediction} on path-rich KGs, all while utilising significantly fewer parameters and training on consumer-grade hardware.
Our contributions are threefold:

\begin{itemize}
    \item We introduce PathE, a fully entity-agnostic, path-based KGE method requiring $<25$\% of the parameters of current parameter-efficient methods.
    \item We conduct comprehensive experiments, demonstrating PathE's efficiency and competitive performance, in path-rich graphs (FB15k-237, CodeX-Large).
    \item We provide ablation studies, validating our modelling choices and analysing PathE's behaviour with varying path quantities and lengths.
\end{itemize}

\section{Related Work}\label{sec:related-work}

\subsection{Knowledge Graph Embeddings}
Several methods have been developed to perform link prediction and other KG related tasks.
These can be divided into logical rule mining~\cite{AMIE3,SAFRAN,rulen,AnyBURL}, path-based reasoning~\cite{MINERVA,Mwalk,DeepPath}, meta-relational learning~\cite{GMatching,Meta_KGR,MetaR} and KGE methods~\cite{TransE,Rotate,HolE,CompleX,ConvE}.
Rule mining and path-based reasoning methods suffer from poor scalability, given that the number of rules and paths increases exponentially with the size of the graph; while meta-relational methods focus on the task of performing predictions on previously unseen nodes.

KGE methods have become the most prominent, as they produce the best performance and are often used as input to other ML models~\cite{review_kg_acquisition_representation_applications,LLM_KG}

The main limitation of KGE methods lies in their reliance on entity embedding tables, leading to two major drawbacks: embedding table size increases with KG growth, making these methods impractical for \textit{large-scale}, real-world KGs (e.g. Wikidata currently counts 11K+ relation types and 108M+ entities); and new entities require full model retraining, hindering dynamic adaptation (\textit{inductiveness}).

\begin{figure*}[t]
    \centering
    \includegraphics[width=\linewidth]{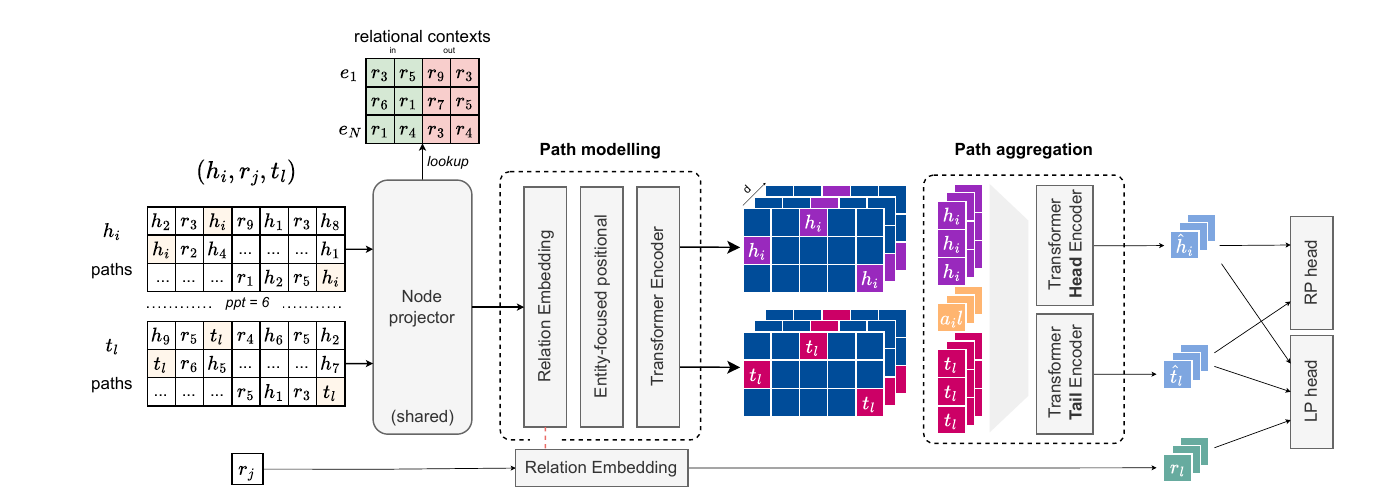}
    \caption{Schematic overview of a PathE architecture, using an example triple $(h_i, r_j, t_l)$ with a set of 6 paths ($ppt$ = 6). The node projector is shared by both head and tail paths, and the relation embedding layer is shared for path modelling and relation $r_j$ encoding. Entity embeddings are aggregated and passed to either the link prediction (LP) or relation prediction (RP) head.}
    \label{fig:pathe-architecture}
\end{figure*}

\subsection{Parameter Efficient Representations} 
Recent work~\cite{nodepiece,EARL} has focused on reducing the amount of stored information by encoding a subset of entities, thus finding a balance between memory requirement and performance.
Nodepiece~\cite{nodepiece} embeds entities as a function of their shortest path distance to the (pre-stored) \textit{anchor} embeddings and their relational context.
Although this method is more efficient than traditional embedding methods and is inductive, it still allocates and learns an embedding table of anchors which increases in proportion to the size of the KG.
Instead, EARL~\cite{EARL} uses relations along with a fixed vocabulary of entity embeddings, called \textit{reserved entities}, and the similarity between relational contexts of reserved entities and every other entity to compute entity representations using a GNN model.
This approach has the ability to inductively embed unseen nodes, achieves better parameter efficiency and outperforms Nodepiece.

Methods based on GNNs typically require storing the whole adjacency matrix, which limits their applicability to larger KGs.
Recently, methods like RED-GNN \cite{REDGNN} have improved over traditional GNN for parameter efficiency.
However, the authors acknowledge computational issues, such as increasing the number of layers in the GNN, which has been observed to limit the scalability of the model \cite{MulGA}.
The scalability of GraIL \cite{GRAIL} for link prediction in standard datasets and its sole evaluation on the inductive setting have also been observed in \cite{REDGNN,NBFnets}.
Similarly, SNRI \cite{SNRI} faces computational issues due to mining of subgraphs between the head and tail of triples, and is thus limited to the inductive setting \cite{MulGA}.
Additionally, NBFNet \cite{NBFnets} and A*Net \cite{A*NEt} achieve very good performance in link prediction by limiting the propagation of messages only between paths connecting the head and tail nodes in a triple.
Despite their performance, the latter methods do not produce individual node representations.
Instead, they only perform link prediction, as the representations of nodes are conditioned on the source node where the message passing begins and they cannot be easily disentangled.

\subsection{Path-based Embedding Methods}
Significant work has focused on using KG-mined paths to leverage their multi-step semantics for link and relation prediction.

These include PTransE~\cite{PTransE} which leverages paths up to length 3 and introduces the path constrained resource allocation algorithm to measure their reliability; PaSKoGE~\cite{Paskoge} which builds upon PTransE and proposes an automated way of calculating the margin hyper-parameter for the loss function; DPTransE~\cite{DPtransE} which extends PTransE by using clustering to group relation types and calculate the weights of paths, while using relation-group specific classifiers to score triples.
Furthermore,  ~\cite{toutanova_compositional},~\cite{relational_path_emb} and ~\cite{path_rotate} all use composition operators to combine path elements into a single representation,~\cite{PRRL,RPJE} and~\cite{InterERP} first mine logical rules which they convert to paths and use them in conjunction with traditional embeddings to perform link prediction; while~\cite{RNN_paths} and~\cite{TransP_entity_path_rela_path_lstm} use recurrent models to combine path elements into a single representation.
Finally,~\cite{pathcon} developed PathCon which utilises relational paths combined with a GNN to perform relation prediction and achieves state-of-the-art performance on the task.
Overall, all of those methods either use paths in conjunction with non-scalable embedding methods~\cite{PTransE,DPtransE,path_rotate}, or mine paths between the head and the tail of each triple~\cite{PTransE,pathcon}.

This path mining process is inherently complex and can become intractable in larger KGs due to the sheer number of potential paths.
Additionally, the resulting entity embeddings are contextualised to specific triples, necessitating the computation of all possible representations for entities across triples.
This poses a challenge when, for example, discovering new triples.

\section{Learning Entity-Agnostic KG embeddings}\label{ch:method}

Given a set $\mathcal{E}$ of entities and a set $\mathcal{R}$ of relations, a Knowledge Graph $\mathcal{K} \subseteq (\mathcal{E} \times \mathcal{R} \times \mathcal{E})$ is a directed multi-relational graph that stores domain knowledge in the form of triples, which are also called facts~\cite{review_kg_acquisition_representation_applications}.
Each triple $(h,r,t)$ consists of a head entity $h \in \mathcal{E}$, a tail entity $t \in \mathcal{E}$ and the relationship $r \in \mathcal{R}$ between those entities.
We denote the number of triples in a batch as $Z$, the number of paths used to describe an entity as $ppe = ppt / 2$ (where $ppt$ stands for paths-per-triple), and the size of the longest path in the batch as $plen$.

\subsection{Path Generation and Representation}\label{sec:pathgen}

Training paths are created by mining random walks from each entity in the KG.
For each entity, we attempt to mine $N$ unique entity-relation paths with no loops (no nodes in the path appearing more than once).
These paths are either \textit{outgoing} (starting from the node) or \textit{incoming} (ending at the node) with equal probability.

Mining paths for each entity in isolation allows for parallelisation, and can easily scale to large KGs.

These paths provide information about the neighbourhoods of the entities and are expected to localise them within the KG.
Moreover, batching together multiple paths for each entity allows the model to extract information related to the different semantics of an entity occurring in the various paths.
For example, the path $<$\textit{Arnold Schwarzenegger, actor in, the Terminator, directed by, James Cameron}$>$ and the path $<$\textit{Arnold Schwarzenegger, winner of, Mister Olympia, year, 1980}$>$ provide very different information for the entity Arnold Schwarzenegger, with each possibly being less or more useful depending on the task at hand.

\subsection{Model Architecture}\label{sec:method}

PathE consists of four modules which are trainable end-to-end: the \textit{node projector}, which maps entities into a continuous space based on their relational context; the \textit{path sequence model}, which processes batches of entity-relation path sequences and produces contextual representations of entities; the \textit{aggregator}, which aggregates the node representations from different paths into a single contextualised representation; and, finally, the \textit{prediction heads}, which produce a score for each triple or relation depending on the task (we provide separate heads for relation and link prediction).
Our model is illustrated in Figure~\ref{fig:pathe-architecture} and described as follows.

\subsection{Node Projector}\label{ssec:projector}

For the node projector, we utilise a two-layer MLP which takes as input the local adjacency matrix of the node-edge graph.
The node-edge graph is a weighted graph created by converting the edges to nodes and adding a directed relation from each edge to a node with weight equal to the number of times this edge appears in the relational context of the node.

We utilise a separate projector for the incoming and outgoing contexts.
The input of the MLP is the adjacency matrix $\mathbf{A}_{er}$ and the output is a matrix $\mathbf{P} \in \mathbb{R}^{|\mathcal{E}| \times d}$ resulting from an affine transformation with non-linear activation:
 \begin{equation}
    \mathbf{P} = \mathbf{W}_{2}Relu(\mathbf{W}_{1}\mathbf{A}_{er} + \mathbf{b}_1) + \mathbf{b}_2.
\end{equation}

The representations produced for the incoming $\mathbf{P}_{in}$ and outgoing $\mathbf{P}_{out}$ contexts are then fused together through a two-layer MLP and projected to $\mathbf{P} \in \mathbb{R}^{|\mathcal{E}| \times d}$ as follows:
\begin{equation}\label{eq:proj-inout}
     \mathbf{P} = \mathbf{W}_{2}Relu(\mathbf{W}_{1}(\mathbf{P}_{in} | \mathbf{P}_{out}) + \mathbf{b}_1) + \mathbf{b}_2,
\end{equation}
where $|$ denotes the concatenation operator among tensors, and $d$ is the embedding dimension.

\subsection{Path Modelling}\label{ssec:pathe-path-modelling}

Path modelling is operated via a self-attention layer (specifically, a Transformer encoder) on a batch of projected paths $\mathbf{B} \in \mathbb{R}^{paths \times plen \times d}$, where $paths = Z \times ppe \times 2$.
At this stage, each path consists of nodes and edges which are projected into a continuous space $\mathbb{R}^d$ using the node projector and an embedding layer for relations, respectively.

To help the model localising the head and tail entities within the paths, a learned positional encoding is added to each embedding vector in the path, based on its position in the sequence.
For instance, for an entity path $e_0, \dots, e_9$ where $e_i \in \mathcal{E}$ where the head appears as the 5th element, we consider the following positional encodings for this path

$$e_0, e_1, e_2, e_3, \mathbf{e_4}, e_5, e_6, e_7, e_8, e_9$$
$$[5, \ 4,\ \ 3,\ \ 2,\ \ 1,\ \ 2,\ \ 3,\ \ 4,\ \ 5,\ 6]$$

and add the corresponding embedding of each position (instead of using traditional positionals $[1, 2, \dots, 10]$).
These \textit{entity-focused positional encodings} can be seen as a path-level contextualisation of \cite{devlin2018bert}, which in turn improves the cyclical positional encoding of \cite{attention}.
This also comes with the advantage of potentially learning to discount the contribution/relevance of entities that are far from the head or tail.
Paths are then passed through the Transformer encoder, which attends to all the elements of each sequence and produces the path-contextualised projections (no masking is necessary).

\subsection{Path Aggregator}\label{ssec:pathe-aggregation}

As shown in Figure~\ref{fig:pathe-architecture}, after paths are passed through the Transformer encoder, the output representations are a tensor of shape $\mathbf{B}_{out} \in \mathbb{R}^{paths \times plen \times d}$, where, for each path, every entity and relation has an embedding vector of size $d$.
From the output, the entity representations of the head and tail are selected from $\mathbf{B}_{out}$, resulting in two tensors denoted as $\mathbf{B}_{head},\mathbf{B}_{tail} \in \mathbb{R}^{Z \times ppe \times d}$.
Each tensor thus contains all the embeddings of the same entity from its $ppe$ paths.
These representations are then aggregated into a single $d$-dimensional representation for each entity.
For this, we experiment with three aggregation strategies: (i) averaging all entity vectors; (ii) using a bidirectional recurrent encoder with LSTM units~\cite{LSTM_ori}; or (iii) using a Transformer encoder.
Of the three approaches, the averaging baseline is the simplest and uses no learned weights to perform the aggregation, while the others learn to aggregate the entity vectors and use separate encoders for head and tail entities.

In contrast, the Transformer aggregator takes as input the sequence of the head embeddings concatenated with the sequence of the tail embeddings, in addition to an aggregation token (denoted as $a_{il}$ in Figure~\ref{fig:pathe-architecture}) which is randomly initialised.
This is expected to aggregate and contextualise the entity embeddings.
The output of the aggregator is a tensor of shape $ \mathbb{R}^{Z \times d}$ for head and tail entities respectively.

\begin{figure}[t]
    \centering
    \includegraphics[width=\linewidth]{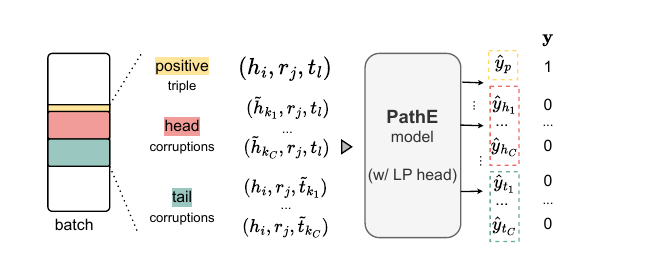}
    \caption{Example portion of a training batch highlighting a positive triple and its stack of head and tail corruptions (negative triples) for training the link prediction (LP) head.}
    \label{fig:pathe-batch}
\end{figure}

\subsection{Training Objective}\label{ssec:pathe-training}

The ability to make the aggregated representation expressive, and capable to be used in KG completion tasks, depends on the training objective.
This is designed on top of the path aggregator, and its formulation currently depends on the type of invariance that representations are expected to have.
In our case, as the goal is to find missing links in the KG, we focus on \textit{relation prediction} and \textit{true triple classification} as a surrogate task for link prediction.
Relation prediction aims at predicting the relation(s) that may exist between head and tail entities -- hence completing the triple $(h, ?, t)$.
Link prediction is of more general scope, as it aims at predicting either the head entity $h$ given the incomplete triple $(?, r, t)$; or analogously, the tail entity $t$ from $(h, r, ?)$.

To accomplish this, we implemented two distinct heads atop the path aggregator as separate training objectives.
Currently, PathE is trained using either of these heads, contingent upon the downstream task.
We leave the investigation of both heads for multi-task learning as future work.

\subsubsection{Relation Prediction Head}\label{sssec:rp-head}
Once the path contextualised representations of the head and tail entities have been obtained, they are concatenated and the resulting matrix $\mathbf{F} \in \mathbb{R}^{Z \times 2d}$ is passed through a linear layer which outputs a score matrix $\mathbf{S} \in \mathbb{R}^{Z \times |\mathcal{R}|}$ with the score of each $(head,tail)$ for each relation.
This yields a probability distribution over all the possible relations in $|\mathcal{E}|$, given the head and tail embeddings.
As this is a multi-class classification task, the relation prediction head uses the Cross Entropy loss between the model's prediction and the true relation.

\begin{align}
      \ell(x, y) &= L = \{l_1,\dots,l_N\}^\top, \\
      \text{s.t.} \quad l_n &= - w_{y_n} \log \frac{\exp(x_{n,y_n})}{\sum_{c=1}^C \exp(x_{n,c})}
\end{align}

Cross Entropy has already been demonstrated to be effective for this task \cite{dogs_tricks}.

\subsubsection{Link Prediction Head}\label{sssec:lp-head}

For link prediction, we train for \textit{true triple classification} as a surrogate task.
This is done by concatenating the head, relation, and tail embeddings in a single tensor of dimension $d \times 3$ representing the whole triple $(h, r, t)$; and stacking a fully connected layer for binary classification.
In other words, the head predicts whether the triple is in the training set (positive triple) or not (negative triple).
Negative triples are constructed by head and tail corruption: the former creates new (negative) triples by replacing $h$ with other entities while keeping relation $r$ and tail $t$ unchanged\footnote{As more triples sharing the same relation and tail may exist in the training set, we always filter the entities in order not to create false negatives (which would also affect the evaluation otherwise).}; whereas tail corruptions are created analogously by fixing $h$, $r$ while changing $t$.
Figure~\ref{fig:pathe-batch} illustrates an example partition of a training batch.

After sampling $N$ head and $N$ tail corruptions, the model is trained to classify each triple as positive or negative.
To balance the classification task, the Binary Cross Entropy loss equally weighs the contribution of positive and negatives.
This is done by dividing the sum of the negative losses by $N \times 2$ as per \cite{NBFnets}.
In line with the literature, we also experiment with Cross Entropy and the self-adversarial negative sampling loss proposed in \cite{Rotate}.

\section{Experiments}\label{sec:experiments}

To evaluate our method while addressing the challenges outlined in the introduction, we focus on the following research questions:
(\textbf{RQ1}, Encodings) To what extent can we learn parameter efficient KG embeddings by only encoding relationships and paths?
(\textbf{RQ2}, Path Learning) How can we best leverage entity-relation paths to encode the KG structure and learn informative representations for link prediction tasks?
and (\textbf{RQ3}, Path Setup) How does the path length and the number of paths per triple influences model performance?

To address RQ1, we train a grid of models on common KG benchmarks and compare performances with baselines and state-of-the-art KGE methods for \textit{transductive link prediction}, \textit{inductive link prediction} and \textit{relation prediction}.

Multiple configurations of PathE with different number of paths and entity aggregation strategies are also tested to trace the contribution of each component related to the use of paths (RQ2).
Finally, we experiment with varying number of paths per entity and visualise the custom positional embeddings to study the influence of the path number and length (RQ3).

\input{tables/linkpred}
\input{tables/relpred}

\subsection{Experimental Setup}\label{ssec:expsetup}

In line with the literature~\cite{nodepiece}, we chose four benchmark datasets, FB15k-237, YAGO3-10, CoDEx-Large and WN18RR to evaluate our model on KGs of various sizes and characteristics (c.f. Table~\ref{tab:dataset-stats}).
FB15k-237 \cite{toutanova2015representing} and CoDEx-Large \cite{safavi2020codex} are derived from Freebase and Wikidata respectively, while WN18RR (from WordNet) and YAGO3-10 focus on more specific domains like lexical relations and person attributes.

We compare our model with both state-of-the-art and parameter efficient KG embedding models, including RotatE \cite{Rotate}, NodePiece \cite{nodepiece} and EARL \cite{EARL}.
We also include a NodePiece model without anchors as it is the only method that, like PathE, is fully entity agnostic (it does not encode any anchors/reserved entities) and maintains parameter efficiency and inductiveness.

\textbf{Evaluation.} All models are evaluated on link and relation prediction, using the original train, validation, and test splits.
We report Mean Reciprocal Rank (MRR) and Hits@K in the filtered setting \cite{TransE}.
Both these metrics are computed using the scores of the triples produced by the model and evaluated by the ranking induced from those scores.
Hits@K measures the ratio of true triples that are ranked among the top K, whereas the MRR averages the reciprocal ranks of true triples and drops rapidly as the ranks grow.
These measures are computed by sampling $N \times 2$ corruptions (negative triples) for each positive: N negatives for head corruptions, by replacing the head with other entities in $|\mathcal{E}|$; and N negatives for tail corruptions, which is analogous to the former case. In our experiments, we use the full set of entities $|\mathcal{E}|$ to produce corruption for the evaluation of our model. 
Furthermore, we reuse the \texttt{Effi} metric proposed in \cite{EARL} to quantify the efficiency of models as performance cost.
This is calculated as $MRR/M(P)$, where $M(P)$ denotes the number of trainable parameters.
For all the compared models, we report parameter count and prediction metrics from their corresponding articles.

\textbf{Implementation.} Our model is implemented in PyTorch v2.1 \cite{pytorch} using PyTorch Lightning 2.1 and PyKEEN v1.1 \cite{pykeen}.
Experiments were run on an Intel Core i9-13900 with 128GB RAM and an NVIDIA RTX 3090 GPU.
All models are trained with an early stopping criterion (patience at 10, min delta at 0) and use 99 negative triples for validation.
The code can be found at \url{https://github.com/IReklos/PathE}.

\subsection{Transductive Link Prediction}\label{ssec:link-prediction}

We trained a grid of PathE models with the Link Prediction head, computing MRR and Hits@K by ranking each true (test) triple against all its head and tail corruptions.
Optimal models were found via random search, sampling 50 configurations from a search space for FB15k-237 and WN18RR due to their size.
The best hyper-parameters for CoDEx-Large and YAGO3-10 were derived from these results.

Results are given in Table~\ref{tab:lp-results} for all benchmarks.
Overall, our model outperforms Nodepiece w/o anchors on all benchmarks and achieves competitive performance to Nodepiece (with anchors) on FB15k-237 and CoDEx-Large, while requiring less 25\% of the parameters.
More precisely, the MRR on FB15k-237 is only 0.04 less than Nodepiece with anchors and 0.012 more than Nodepiece w/o anchors while using less than 10\% and less than 25\% of the parameters respectively; which confirm PathE as the most efficient model ($Effi = 1.03$) compared to EARL ($Effi = 0.17$) and NodePiece without anchors ($Effi = 0.15$).
On CoDEx-Large, our model outperforms Nodepiece w/o anchors in MRR by 0.081 and only has a deficit of 0.046 compared to Nodepiece with anchors; thus recording $Effi = 0.21$ compared to $Effi = 0.10$ for EARL and $Effi = 0.105$ for Nodepiece w/o anchors.
On WN18RR and YAGO3-10, PathE's performance lags behind Nodepiece and EARL. We hypothesise this is attributed to the datasets' characteristics, specifically the limited number of distinct relations (11 and 37, respectively).
This scarcity hinders the unique encoding of KG nodes, affecting the model's ability to differentiate between them.
Despite this limitation, PathE demonstrates superior performance to Nodepiece w/o anchors on MRR across both datasets.
Notably, PathE achieves over 6 times the performance on WN18RR, with an efficiency of $0.10$ compared to $0.04$, and more than $2\times$ the performance on YAGO3-10, reaching an $Effi$ of $0.25$ compared to $0.05$.

Despite the size of CoDEx-Large ($2\times$ more training triples and $5\times$ more entities than FB15k-237) we recall that the parameter budget of our model scales linearly with the number of relations (69 for CoDEx, 237 for FB15k).
The nearly tripling of the parameter count of the model trained on CoDEx-Large is due to the use of embeddings of dimensionality $d=128$ instead of $d=64$ for FB15k-237 and the use of two encoder layers in the path modelling transformer and the aggregator module.
Instead, Nodepiece and EARL are affected by the number of entities, as they both allocate embedding tables for a subset of them.

Details on the hyper-parameter settings and best configurations are provided in Appendix~\ref{ssec:appendix-lp}.
The results of the \textit{inductive link prediction} experiments are presented in Appendix~\ref{ssec:appendix-lp-inductive}.

\subsection{Relation Prediction}\label{ssec:relation-prediction}

To evaluate the representations of our model for relation prediction, we train and evaluate with the associated head.
This is only done for FB15k-237 and WN18RR, as these are the only benchmarks where a parameter-efficient method has been evaluated and reported in \cite{nodepiece}.

Relation prediction results are outlined in Table~\ref{tab:rp-results}.
Our model significantly outperforms both parameter-efficient (Nodepiece) and state-of-the-art (RotatE) models on relation prediction, while requiring less than a million parameters for FB15k-237.
Notably, the PathE model for WN18RR only has 40K parameters.
For this dataset, we found that an embedding dimension of 32 is sufficient for relation prediction; as increasing the embedding dimension did not bring about any increases in performance.
This is substantially lower compared to the other models (2-3 orders of magnitude).
Details on the hyper-parameter settings are given in Appendix~\ref{ssec:appendix-rel-pred}.

\subsection{Ablation Study}\label{ssec:ablation}

\input{tables/ablation}

To quantify the contribution of each component in the model, we carried out ablation studies on the best performing benchmarks for link prediction: FB15k-237 and CoDEx-Large.
The ablation dimensions are summarised as follows.

\begin{itemize}
    \item w/o Aggregator, by replacing the Transformer Encoder with a simple averaging operation over the entity embeddings of different paths.
    \item w/o Multiple paths, where we use only 1 path per entity in each triple (1 for the head, 1 for the tail), sampled randomly. Hence, no aggregation is necessary.
    \item w/o Entity-focused positional encodings, where positional information is injected by adding the relative position of each element in the path, regardless of where the head/tail occurs within the path.
\end{itemize}

The ablation results are summarised in Table~\ref{tab:ablation}.
Overall, we found the averaging operator to perform slightly worse than the transformer aggregator ($MRR=0.215$ instead of $MRR=0.216$ for FB15K-237, and $MRR=0.132$ instead of $MRR=0.144$ for CoDEx-Large); although the difference is very small on FB15k-237 where the averaging produces better performance in Hits@5 and Hits@10.

On both datasets, our results confirm the contribution of using multiple paths per head and tail, rather than a single sequence per entity ($MRR=0.216$ to $MRR=0.184$ for FB15k-237; and $MRR=0.144$ to $MRR=0.105$ for CoDEx-Large).

Finally, entity-focused positional encodings improve performance by 13\% on FB15k-237, while achieving a 22\% gain on CoDEx-Large.

\begin{figure}[t]
    \centering
    \includegraphics[width=\linewidth]{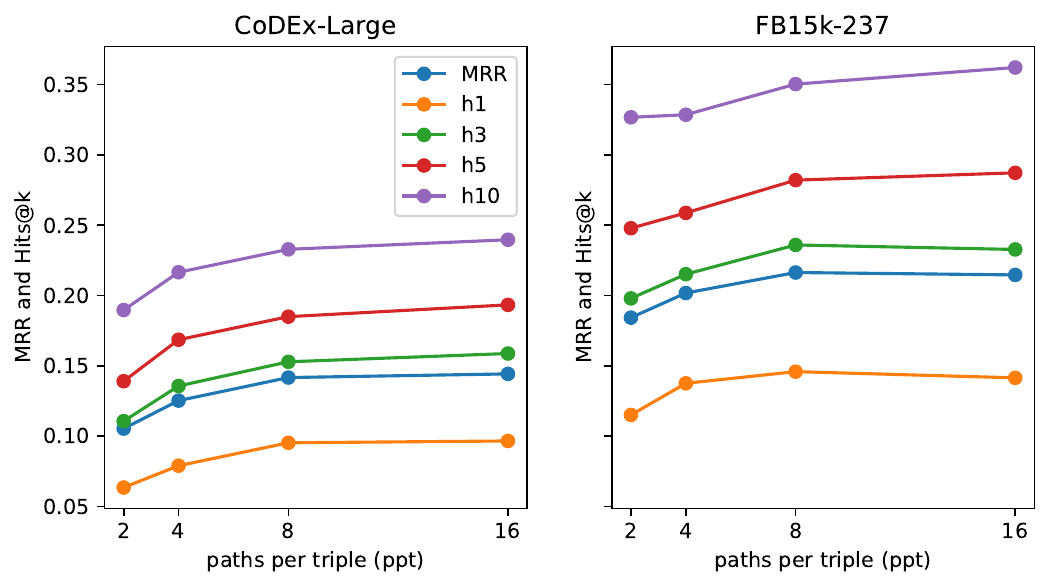}
    \caption{Link prediction performance of PathE (MRR and Hits@K), in relation to the number of paths per triple.}
    \label{fig:pathe-ppe}
\end{figure}

\begin{figure}
    \centering
    \includegraphics[width=.95\linewidth]{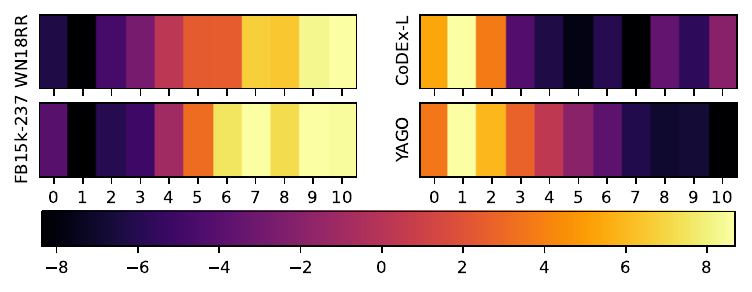}
    \caption{Visualisation of the relative positional embeddings for entities after dimensionality reduction via PCA (1 dim).}
    \label{fig:pathe-plength}
\end{figure}

\subsection{Path Number and Length}

We investigated the effect of varying the number of paths per triple (\textit{ppt}) on model performance. Using the best-performing model on FB15k-237 and CoDEx-Large, we tested with $2, 4, 8$, and $16$ paths per triple (equivalent to $1, 2, 4, 8$ paths per entity).
Figure~\ref{fig:pathe-ppe} shows that increasing the number of paths generally improves MRR, but gains plateau.
For CoDEx-Large, performance leveled off after 8 paths per entity, with a minimal MRR increase of $0.0026$ from $4$ to $8$ paths.
For FB15k-237, performance plateaued after $4$ paths per entity, with a decrease of $0.0017$ beyond that point.

While experiments used paths of length 20, entity-focused positionals were found to improve performance and help the model discount entities further from the head or tail.
Figure~\ref{fig:pathe-plength} shows a PCA visualisation of embedding activations, revealing a separation between lower (closer to head/tail) and higher positions.

From a manual inspection, we can see that the embeddings of lower positions (those that are closer to the head or tail entity) appear separated from embeddings of higher positions, with a reversal in the magnitude of the activations.
In CoDEx-Large the separation happens at position 2, hence the model focuses on paths of length 7, while in FB15k-237 the separation happens in position 4 which shows that the model focuses on paths of length 11.

\subsection{Scope and Applicability of PathE}\label{sec:scope-limitations}

PathE demonstrates its effectiveness as a path-based KGE method, particularly on densely connected KGs with high relational diversity.
As detailed in Appendix~\ref{sec:suitability}, PathE leverages rich relational contexts to generate discriminative entity embeddings, making it well-suited for tasks such as link prediction and relation prediction.
Its scalability and parameter efficiency further enhance its reuse as a lightweight model.

PathE's performance still depends on certain KG properties.
It is less effective on sparser KGs with lower relational diversity, average entity degree, and unique relational contexts, such as WN18RR and YAGO3-10.
We posit these characteristics constrain the model's ability to differentiate between entities and learn expressive embeddings. In such cases, alternative methods like \cite{nodepiece} or \cite{EARL}, which rely on anchors or reserved entities, may be more suitable.
Despite this, PathE remains a state-of-the-art solution for relation prediction, and its ability to capitalise on relational richness and path diversity underscores its value for large-scale KGs, where these characteristics are prevalent.

\section{Conclusion}\label{sec:discussion}

This work introduces PathE, an entity-agnostic KG embedding method that dynamically computes entity embeddings by aggregating path information. PathE eliminates the need for pre-allocated embedding tables, requiring less than 25\% of the parameters used by existing lightweight methods, with memory usage scaling linearly with the relation vocabulary.

Experiments on link prediction demonstrate PathE's effectiveness on datasets like FB15k-237 and CoDEx-Large, characterized by high relational diversity and rich path information.
Ablation studies also highlight the added value of multiple paths in achieving robust performance.
While less competitive on sparser datasets like WN18RR and YAGO3-10, where relational diversity and contextual richness are limited, PathE achieves state-of-the-art performance in relation prediction, surpassing other parameter-efficient methods.

Our primary contribution is in exploring the capacity of path-based contextualisation to learn entity-agnostic embeddings using only relational contexts.
PathE offers a competitive and scalable solution, particularly suited for densely connected KGs with diverse relational vocabularies, where parameter efficiency and a lightweight model are essential for reuse in both link and relation prediction tasks.
Future work will focus on evaluating PathE on larger datasets like Wikidata5M and incorporating multi-task learning to further enhance its adaptability and scalability for web-scale KGs.

\bibliographystyle{named}
\bibliography{ijcai24.bib}

\input{appendix}

\end{document}

%% file: tables/linkpred.tex
{\renewcommand{\arraystretch}{1.2}
\begin{table*}[t]
\centering
\resizebox{\textwidth}{!}{%
\begin{tabular}{lcccccccccccc}
                   & \multicolumn{3}{c}{\textbf{FB15k-237}}          & \multicolumn{3}{c}{\textbf{WN18RR}} & \multicolumn{3}{c}{\textbf{YAGO3-10}} & \multicolumn{3}{c}{\textbf{CoDEx-Large}}        \\ \cline{2-13} 
                   & \#P(M)          & MRR            & Hits@10        & \#P(M)      & MRR       & Hits@10     & \#P(M)       & MRR        & Hits@10     & \#P(M)          & MRR            & Hits@10        \\ \hline
RotatE             & 29 .3         & 0.336          & 0.532          & 40.6      & 0.508     & 0.612       & 123.2      & 0.495      & 0.670       & 78.0          & 0.258          & 0.387          \\
EARL               & 1.8           & 0.310          & 0.501          & 3.8       & 0.440     & 0.527       & 3.0        & 0.302      & 0.498       & 2.1           & 0.238          & 0.390          \\
Nodepiece + Rotate & 3.2           & 0.256          & 0.420          & 4.4       & 0.403     & 0.515       & 4.1        & 0.247      & 0.488       & 3.6           & 0.190          & 0.313          \\ \hline
Nodepiece w/o anchors               & 1.4           & 0.204          & \textbf{0.355}          & \textbf{0.3}       & 0.011     & 0.019       & 0.5        & 0.025      & 0.041       & \textbf{0.6}           & 0.063          & 0.121          \\
\textbf{PathE} & \textbf{0.21} & \textbf{0.216} & \textbf{0.350} & 0.67     & \textbf{0.069}     & \textbf{0.124}        & \textbf{0.24}       & \textbf{0.060}      & \textbf{0.093}       & 0.68 & \textbf{0.144} & \textbf{0.240} \\ \hline
\end{tabular}%
}
\caption{Transductive link prediction results. Parameter count (in millions of parameters), MRR, and Hits@10 for the other models are taken from \protect\cite{EARL}, \protect\cite{nodepiece}. Results are highlighted for fully entity-agnostic models (no anchors/ reserved entities).}
\label{tab:lp-results}
\end{table*}
}

%% file: tables/relpred.tex
{\renewcommand{\arraystretch}{1.3}
\begin{table}[t]
\resizebox{\linewidth}{!}{%
\centering
\begin{tabular}{lcccccc}
                   & \multicolumn{3}{c}{\textbf{FB15k-237}} & \multicolumn{3}{c}{\textbf{WN18RR}} \\ \cline{2-7} 
                   & \#P(M)   & MRR      & Hits@10   & \#P(M)   & MRR    & Hits@10  \\ \hline
RotatE             & 29     & 0.905    & 0.979     & 41     & 0.774  & 0.897    \\
Nodepiece + Rotate & 3.2    & 0.874    & 0.971     & 4.4    & 0.761  & 0.985    \\
\textbf{PathE} & \textbf{0.86}   & \textbf{0.972}   & \textbf{0.998}      & \textbf{0.05}   & \textbf{0.874}   & \textbf{0.999} \\ \hline   
\end{tabular}
}
\caption{Relation prediction results. FB15K-237 and WN18RR results for NodePiece and RotatE are taken from \protect\cite{nodepiece}.}
\label{tab:rp-results}
\end{table}
}

%% file: tables/ablation.tex
{\renewcommand{\arraystretch}{1.3}
\begin{table*}[t]
\centering
\resizebox{\textwidth}{!}{%
\begin{tabular}{lcccccccccccc}
                                & \multicolumn{6}{c}{\textbf{FB15k-237}}                                 & \multicolumn{6}{c}{\textbf{CoDEx-Large}}          \\ \cline{2-13} 
                                & \#P(M) & MRR   & Hits@1 & Hits@3 & Hits@5 & \multicolumn{1}{c|}{Hits@10} & \#P(M) & MRR   & Hits@1 & Hits@3 & Hits@5 & Hits@10 \\ \cline{1-13}
\textbf{PathE} (base model)         & 0.21 & 0.216 & 0.146  & 0.236  & 0.282  & \multicolumn{1}{c|}{0.350}  & 0.68 & 0.144 & 0.100  & 0.159  & 0.193  & 0.240  \\
-no aggregator              & 0.19 & 0.215 & 0.141 & 0.235  & 0.284 & \multicolumn{1}{c|}{0.358}   & 0.48 & 0.132 & 0.083  & 0.145  & 0.179  & 0.231   \\
-no multiple paths              & 0.21 & 0.184 & 0.115  & 0.198  & 0.248  & \multicolumn{1}{c|}{0.327}   & 0.68 & 0.105 & 0.063  & 0.111  & 0.139  & 0.190   \\
-no entity-focused positionals & 0.21 & 0.191 & 0.122  & 0.209  & 0.256  & \multicolumn{1}{c|}{0.327}   & 0.68 & 0.118 & 0.071  & 0.129 & 0.161  & 0.210   \\ \hline
\end{tabular}%
}
\caption{Ablation results of PathE on link prediction on FB15K-237 and CoDEx-Large (the best performing datasets for our model), with parameter count, MRR, and detailed Hits@N metrics (N=\{1, 3, 5, 10\}).}
\label{tab:ablation}
\end{table*}
}

%% file: appendix.tex

\clearpage  
\appendix

\section{KG Embedding Datasets}\label{ssec:appendix-datasets}

We provide additional information on the benchmark datasets chosen to evaluate the performance of our model, while Table \ref{tab:dataset-stats} overviews of the number of relations, entities, and train/validation/test splits for each dataset.
These datasets vary in size, domain, and relational structure, offering a comprehensive assessment of our model's performance.

\begin{itemize}
    \item FB15k-237 \cite{toutanova2015representing} is derived from Freebase \cite{bollacker2008freebase} --  a large collaborative KG that is now part of the Google Knowledge Graph and preceded Wikidata (two example of large scale KGs). It counts 15K entities with a vocabulary of 237 relations.
    \item YAGO3-10 \cite{dettmers2018convolutional} is a subset of YAGO3 where entities have a minimum of 10 relations each. It covers attributes of persons such as citizenship and profession, with 120K+ entities and 37 relations.
    \item CoDEx-Large \cite{safavi2020codex} is a subset of Wikidata that was designed to cover more diverse content, and provide a more difficult benchmark for link prediction. It has ~78K entities and uses 69 relations.
    \item WN18RR \cite{dettmers2018convolutional} is a subset of WordNet \cite{miller1995wordnet}, a lexical taxonomy linking words to their synonyms, hyponyms, and meronyms. It counts 40K+ entities and uses a vocabulary of 11 relations.
\end{itemize}

\section{Transductive Link Prediction}\label{ssec:appendix-lp}

We provide more details on the experimental setup and the hyper-parameter configuration of the models reported in our link prediction experiments (c.f. Section~\ref{ssec:link-prediction}).
Table~\ref{tab:ablation-sspace-lp} documents the search space we defined for the random search of hyper-parameters; whereas Table~\ref{tab:path-confs-lp} reports the configuration of the best performing models on each benchmark dataset.
Due to the size of CoDEx-Large and YAGO 3-10, and the availability of computational resources for our experiments, we remark that random search (with N=50 trials) was carried out only on FB15k-237 and WN18RR -- with the best performing models originating from such experiments.

We remark that all Python code is openly available at \url{https://anonymous.4open.science/r/kg_embeddings/README.md}, together with the main instructions to reproduce the experiments reported in this article. 

\input{tables/dataset_stats}

\input{tables/ablation_sspace}
\input{tables/ablaton_sspace_rp}

\input{tables/confs_lp}

\input{tables/confs_rp}

\section{Inductive Link Prediction}\label{ssec:appendix-lp-inductive}

To further assess PathE's capabilities, we evaluate its performance on the inductive link prediction task, following the benchmark setup proposed by \citeauthor{GRAIL}.
Unlike the transductive setting where the model sees all entities during training (c.f. Section~\ref{ssec:link-prediction}), inductive link prediction requires the model to generalize to entirely unseen entities during inference.
In this setup, the training and inference graphs are completely disjoint: validation and testing are performed on a separate graph containing novel entities.
Consequently, the paths used during inference are composed of entity sequences never encountered during training.
The only commonality between the training and inference graphs lies in the shared set of relation types.
This necessitates the model to learn the underlying semantics of relations rather than simply memorising entity-specific patterns, which is crucial for real-world applications where new entities are constantly introduced.

The inductive link prediction benchmarks derived from FB15k-237 and WN18RR each consist of four versions (V1-V4), as shown in Table~\ref{tab:inductive-stats}, exhibiting varying graph properties and difficulty levels.
These versions differ in terms of entity and interaction counts, offering diverse scenarios to examine PathE's adaptability.
For our experiments, we evaluate PathE on all four versions of the FB15k-237 and WN18RR inductive benchmarks.
We sample 50 negative triples for each positive triple in the test set and report the Hits@10 metric, consistent with prior work \cite{nodepiece,GRAIL}.
The hyperparameter configurations used for each dataset are identical to those found optimal in the transductive experiments, allowing us to directly compare the model's performance across both settings.

The results of the inductive experiments are reported in Table~\ref{tab:inductive-results}.
PathE outperforms all other path-based methods in FB15k-237 V1 and V2 and is marginally outperformed by RuleN \cite{rulen} on V3 an V4.
At the same time, PathE outperforms GraIL \cite{GRAIL} (a GNN based method) in V1, V2 and V3 and manages to match the performance of NBFNet \cite{NBFnets} (which is the best performing GNN method overall) on V1.


Table~\ref{tab:inductive-results} reports the results of the inductive experiments.

\textbf{FB15k-237}. PathE demonstrates promising generalisation capabilities on the FB15k-237 inductive benchmark.
It outperforms all other path-based methods on versions V1 and V2, highlighting its ability to effectively leverage relational paths for reasoning about unseen entities.
Moreover, PathE surpasses the performance of the GNN-based method GraIL \cite{GRAIL} on V1, V2, and V3, and achieves performance on par with NBFNet \cite{NBFnets} on V1.
While RuleN \cite{rulen} achieves slightly better results on V3 and V4, PathE's strong performance across multiple versions of FB15k-237 underscores its potential for inductive link prediction in KGs with rich relational structure.

\textbf{WN18RR}. On its inductive benchmark, PathE's performance, while lower than some other methods, reveals valuable insights.
As observed in the transductive experiments (Table \ref{tab:lp-results}), WN18RR's high sparsity and limited relational diversity (only 11 unique relations) pose significant challenges for models like PathE that rely heavily on relational contexts.
Despite this, the model exhibits consistent performance across V1, V2, V3, and V4, unlike other path-based methods that show considerable performance variability. 
Notably, PathE outperforms all other path-based methods on V3, the version with the highest number of unique relations among the WN18RR subsets.
This supports our hypothesis that increased relational diversity allows PathE to better differentiate between nodes, even in inductive settings.
These findings corroborate the observations made by \citeauthor{nodepiece} regarding the limitations of relationally-impoverished datasets for methods that depend primarily on relations and their contexts.

\input{tables/indlinkpred}
\input{tables/inductive_properties}

\section{Relation Prediction}\label{ssec:appendix-rel-pred}
Similarly to the transductive link prediction experiments, Tables \ref{tab:ablation-sspace-rp} and \ref{tab:path-confs-rp} report the hyper-parameter search space and the configuration of the best performing models for relation prediction, respectively.
Results are reported for FB15k-237 and WN18RR to ensure comparability (c.f. Section~\ref{ssec:relation-prediction}).


\begin{table*}[t]
\begin{minipage}{.44\linewidth}
    \centering
    \begin{tabular}{ccc}
        \multicolumn{3}{c}{\textbf{WN18RR}}                          \\
        \textbf{Edge ID} & \textbf{Frequency} & \textbf{\% of Total} \\ \hline
        3                & 34796              & 40.1                \\
        1                & 29715              & 34.2        \\
        5                & 7402               & 8.5                  \\
        2                & 4816               & 5.5                  \\
        9                & 3116               & 3.6                  \\
        4                & 2921               & 3.4                  \\ 
        0                & 1299               & 1.5                  \\
        10               & 1138               & 1.3                  \\
        6                & 923                & 1.1                  \\
        7                & 629                & 0.7                  \\
        8                & 80                 & 0.1                  \\\hline
    \end{tabular}
\end{minipage}\hfill 
\begin{minipage}{.44\linewidth}
    \centering
    \begin{tabular}{ccc}
        \multicolumn{3}{c}{\textbf{YAGO 3-10}}                       \\
        \textbf{Edge ID} & \textbf{Frequency} & \textbf{\% of Total} \\ \hline
        21               & 373783             & 34.64                \\
        33               & 321024             & 29.75                \\
        27               & 88672              & 8.21                 \\
        13               & 66163              & 6.13                 \\
        34               & 44978              & 4.16                 \\
        0                & 32155              & 2.98                 \\
        23               & 32055              & 2.97                 \\
        18               & 24068              & 2.23                 \\
        20               & 10710              & 0.99                 \\
        3                & 9248               & 0.85                 \\
        14               & 7754               & 0.71                \\\hline
    \end{tabular}
  \end{minipage}
\caption{Absolute and relative counts of relation occurrences in WN18RR and YAGO 3-10.}
\label{tab:edge-stats}
\end{table*}

\begin{table*}[h]
\begin{minipage}{.44\linewidth}
    \centering
    \begin{tabular}{ccc}
        \multicolumn{3}{c}{\textbf{CoDEx-Large}}                          \\
        \textbf{Edge ID} & \textbf{Frequency} & \textbf{\% of Total} \\ \hline
        4                & 169091              & 30.7                \\
        36                & 60262              & 10.9        \\
        26                & 35979             & 6.5                \\
        17                & 34372               & 6.2                  \\
        62                & 30318               & 5.5                  \\
        14               & 24352               & 4.4                  \\ 
        21               & 23171               & 4.2                  \\
        29               & 22132               & 4.0                  \\
        50               & 17057                & 3.1                  \\
        5               & 11999                & 2.2                 \\
        1               & 11453                 & 2.1                  \\\hline

    \end{tabular}
\end{minipage}\hfill 
\begin{minipage}{.44\linewidth}
    \centering
    \begin{tabular}{ccc}
        \multicolumn{3}{c}{\textbf{FB15k-237}}                       \\
        \textbf{Edge ID} & \textbf{Frequency} & \textbf{\% of Total} \\ \hline
        10               & 15989           & 5.9                \\
        96               & 12893           & 4.7                \\
        9                & 12157           & 4.5                 \\
        191              & 10945           & 4.0                 \\
        68               & 9494            & 3.5                 \\
        3                & 9465            & 3.5                 \\
        12               & 8423            & 3.1                 \\
        85               & 7268            & 2.7                 \\
        11               & 6277            & 2.3                 \\
        149              & 5880            & 2.2                 \\
        4                & 5673            & 2.1                 \\\hline
    \end{tabular}
  \end{minipage}
\caption{Absolute and relative counts of the most frequent relation occurrences in CoDEx-Large and FB15k-237.}
\label{tab:edge-stats-2}
\end{table*}

\begin{table}[h]
    \centering
    \begin{tabular}{ccc}
        \multicolumn{3}{c}{\textbf{Wikidata}}                          \\
        \textbf{Edge ID} & \textbf{Frequency} & \textbf{\% of Total} \\ \hline
        P2860                & 304466243              & 18.6                \\
        P1545                & 199991550              & 12.2        \\
        P2093                & 149677775             & 9.1                \\
        P31                & 121504159               & 7.4                  \\
        P813                & 109489889               & 6.7                 \\
        P248               & 108457782               &6.6                 \\ 
        P854               & 82353706               & 5.0                 \\
        P698               & 68122905               & 4.2                 \\
        P1476               & 55678742                & 3.4                  \\
        P577               & 55247544                & 3.4                \\
        P1433               & 45803735                 & 2.8                  \\\hline

    \end{tabular}
\caption{Absolute and relative counts of the most frequent relation occurrences in Wikidata. Data from \protect\footnotemark[1] and \protect\footnotemark[2].}
\label{tab:edge-stats-3}
\end{table}

\section{Analysis of KGs Properties and their Influence on PathE Performance}\label{sec:suitability}

PathE's performance is intrinsically linked to the structural properties of the underlying KGs.
Specifically, the model relies on encoding entities based on their relational contexts and the paths traversing those relations.
The richness and diversity of these contexts are thus central for generating discriminative entity representations that are effective for link prediction.
This section examines the relationship between the structural characteristics of KGs and the efficacy of PathE.
The experiments detailed in Section~\ref{sec:experiments} demonstrate that PathE performs well on FB15k-237 and CoDEx-Large but yields less competitive results on WN18RR and YAGO3-10.
We posit that these performance differences are primarily attributable to variations in relational diversity, entity degree, and the uniqueness of relational contexts.

\begin{enumerate}
    \item \textbf{Relational Diversity.} A greater number of distinct relation types provides a more expressive vocabulary for characterising entities and their interconnections. This potentially facilitates the model's ability to capture more nuanced relationships and distinguish between entities with greater accuracy.
    \item \textbf{Average Entity Degree.} A higher average degree generally signifies a denser graph with more connections per entity. This translates to a greater number of paths traversing each entity, providing the model with more contextual information for learning entity embedding.
    \item \textbf{Uniqueness of Relational Contexts.} The extent to which entities can be uniquely identified based solely on their relational contexts is also crucial. A higher proportion of unique contexts indicates that the relational structure provides strong discriminatory signals for differentiating between entities.
\end{enumerate}

Tables \ref{tab:dataset-stats}, \ref{tab:edge-stats} and \ref{tab:edge-stats-2} provide insights into the structural differences between the benchmark datasets. We analyse these differences in relation to PathE's performance.

\begin{itemize}
    \item \textbf{WN18RR and YAGO3-10 (sparser KGs).} These datasets are characterized by limited relational diversity, possessing only 11 and 37 distinct relation types, respectively. Furthermore, they exhibit a highly skewed distribution of relations. In WN18RR, the two most frequent relations account for over 74\% of all triples, while in YAGO3-10, they constitute over 64\%. This combination of limited relational diversity, low average entity degree (4.28 for WN18RR and 17.52 for YAGO3-10), and a paucity of unique relational contexts (8\% for WN18RR and 19.8\% for YAGO3-10) significantly hinders PathE's ability to learn discriminative entity embeddings. The relational contexts are simply too homogeneous to effectively distinguish between entities. The results of the inductive experiments (Section~\ref{ssec:appendix-lp-inductive}), where PathE performs poorly on WN18RR V1, V2, and V4, further corroborate this observation. These particular datasets have the lowest number of unique relations (9, 10, and 9, respectively), which constrains the model's capacity to learn the underlying semantics of the relations.

    \item \textbf{FB15k-237 and CoDEx-Large (path-rich KGs):} In contrast, these KGs exhibit considerably higher relational diversity, with 237 and 69 distinct relation types, respectively, and a more balanced distribution of relations (see Table~\ref{tab:edge-stats-2}). Their higher average entity degrees (37.5 for FB15k-237 and 14.14 for CoDEx-Large) contribute to a richer set of paths for each entity. Crucially, they also have a much larger proportion of unique relational contexts (92\% for FB15k-237 and 46.4\% for CoDEx-Large).
    These factors enable the model to effectively capture the diverse relationships and generate more discriminative entity representations, resulting in superior link prediction performance. The inductive experiments, where PathE outperforms other path-based methods on FB15K-237 V1 and V2, further support this claim. Additionally, PathE matches the performance of the best-performing method on FB15K-237 V1.
\end{itemize}

\footnotetext[1]{Wikidata Datamodel Statements. Online: Grafana Dashboard. Accessed on 2025-01-20. URL: \url{https://grafana.wikimedia.org/d/000000175/wikidata-datamodel-statements?orgId=1\&refresh=30m}}

\footnotetext[2]{Wikidata:Database reports/List of properties/Top100. Online: Wikidata. Accessed on 2025-01-20. URL: \url{https://www.wikidata.org/wiki/Wikidata:Database_reports/List_of_properties/Top100}}

This analysis underscores that PathE performs better in KGs characterized by high relational diversity, as well as rich and unique relational contexts.
These characteristics are frequently observed in large, real-world KGs.
For instance, Wikidata counts 12,353 unique relations (called properties), more than 115 million entities (items), and contains 1.6+ billion statements to date\footnotemark[1].
As reported in Table~\ref{tab:edge-stats-3}, Wikidata has a balanced distribution of relations and an average node degree of 29.
While more experiments are needed to assess PathE's applicability to Wikidata, here we remark that these characteristics are better represented by the FB15k-237 and CoDEx-Large benchmarks, which are indeed derived from Freebase and Wikidata, respectively.

In conclusion, while parameter-efficient methods such as NodePiece \cite{nodepiece} and EARL \cite{EARL}, which depend on anchors or reserved entities, are preferable for KGs with limited relational diversity and low uniqueness of relational contexts, PathE emerges as a highly competitive and scalable solution for large, densely connected KGs with diverse relational vocabularies.
This is especially relevant when a lightweight model is required for link prediction.
Nonetheless, and irrespective of the efficiency requirements, PathE achieves state-of-the-art performance in relation prediction, surpassing other parameter-efficient methods and making it a particularly compelling choice when relation prediction is the primary objective.

%% file: tables/dataset_stats.tex
\begin{table}[b]
\centering
\resizebox{\linewidth}{!}{%
\begin{tabular}{@{}llllll@{}}
\toprule
\textbf{Dataset} & \textbf{\#Ent} & \textbf{\#Rel} & \textbf{\#Train} & \textbf{\#Valid} & \textbf{\#Test} \\ \midrule
FB15k-237        & 14,505         & 237            & 272,115          & 17,526           & 20,438          \\
WN18RR           & 40,559         & 11             & 86,835           & 2,824            & 2,924           \\
CoDEx-L          & 77,951         & 69             & 551,193          & 30,622           & 30,622          \\
YAGO3-10         & 123,143        & 37             & 1,079,040        & 4,978            & 4,982           \\ \bottomrule
\end{tabular}
}
\caption{No. of entities, relations, and triples in each dataset.}
\label{tab:dataset-stats}
\end{table}

%% file: tables/ablation_sspace.tex
\begin{table}[h]
\centering
\begin{tabular}{@{}lr@{}}
\toprule
\textbf{Hyper-parameter}                    & \textbf{Range}                       \\ \midrule
Embedding dimension               & \{64, 128, 256\}            \\
\# paths per entity               & \{1, 2, 4,  8\}           \\
Path Transformer                        & \multicolumn{1}{l}{}        \\
\quad Dim feedforward             & \{64, 128, 256\}            \\
\quad \# attention heads          & \{2, 3, 4\}                 \\
\quad \# layers                   & \{2, 3\}                    \\
\quad Dropout                     & \{0.1, 0.2\}                \\
Entity aggregation strategy       & \{avg, trf, LSTM\}                \\
Loss function                     & \{CE, BCE, NSSA\}           \\
\quad Label smoothing             & \{0, 0.1, 0.01\}            \\
\# Negative samples               & \{16, 32, 64, 128\}         \\
Optimiser                         & \{Adam\}                    \\
\quad Learning rate               & \{0.01, 0.001, 0.005\}       \\
\quad Batch size                  & \{64, 128, 256, 512, 1024\} \\
\quad Accumulated batches         & \{16, 32, 64, 128\}         \\ \bottomrule
\end{tabular}
\caption{Hyper-parameter search space for link prediction.}
\label{tab:ablation-sspace-lp}
\end{table}

%% file: tables/ablaton_sspace_rp.tex
\begin{table}[h]
\centering
\begin{tabular}{@{}lr@{}}
\toprule
\textbf{Hyper-parameter}                    & \textbf{Range}                       \\ \midrule
Embedding dimension               & \{32, 64, 128\}            \\
\# paths per entity               & \{4, 8\}           \\
Path Transformer                        & \multicolumn{1}{l}{}        \\
\quad Dim feedforward             & \{64, 128, 256\}            \\
\quad \# attention heads          & \{2, 3, 4\}                 \\
\quad \# layers                   & \{2, 3\}                    \\
\quad Dropout                     & \{0.1, 0.2\}                \\
Entity aggregation strategy       & \{avg\}                \\
Loss function                     & \{CE\}           \\
\quad Label smoothing             & \{0, 0.1, 0.01\}            \\
Optimiser                         & \{Adam\}                    \\
\quad Learning rate               & \{0.01, 0.001, 0.005\}       \\
\quad Batch size                  & \{512, 1024, 2048\} \\
\quad Accumulated batches         & \{8, 16\}   \\\bottomrule
\end{tabular}
\caption{Hyper-parameter search space for relation prediction.}
\label{tab:ablation-sspace-rp}
\end{table}

%% file: tables/confs_lp.tex
\begin{table*}[t]
\centering
\begin{tabular}{@{}lcccc@{}}
\toprule
\textbf{Hyper-parameter}      & \textbf{FB15K-237} & \textbf{WN18RR} & \textbf{CoDEx-L} & \textbf{YAGO3-10} \\ \midrule
Embedding dim        & 64        & 128    & 128     & 64       \\
Paths per entity     & 4         & 2     & 8        & 2        \\
Path Transformer dim       & 256  & 256     & 256     & 256       \\
Path Transformer heads     & 2   & 4      & 2       & 4         \\
Path Transformer layers    & 1   & 2      & 2       & 2         \\
Path Transformer dropout   & 0.1 & 0.1    & 0.1     & 0.1       \\
Entity aggregation   & Transformer Enc       & Transformer Enc    & Transformer Enc     & Transformer Enc       \\
Aggregation layers  &  1          & 2        & 2     & 2  \\
Loss function        & CE        & CE     & CE      & CE        \\
\# Negative samples  & 99        & 99     & 99      & 99        \\
Learning rate        & 1e-3      & 1e-3   & 1e-3    & 1e-3      \\
Batch size           & 4096       & 2048    & 4096    & 1024       \\
Accumulated batches  & 8        & 32      & 16     & 32        \\
Label smoothing      & 0.01         & 0.1    & 0.01       & 0.2      \\ \midrule
Parameter count      & 0.21      & 0.67    & 0.68    & 0.24    \\
Effi = $MRR/P(M)$    & 1.03     & 0.100  & 0.210   & 0.250     \\ \bottomrule
\end{tabular}
\caption{PathE best hyper-parameter configurations for the transductive link prediction experiments.}
\label{tab:path-confs-lp}
\end{table*}

%% file: tables/confs_rp.tex
\begin{table}[t]
\centering
\begin{tabular}{@{}lcccc@{}}
\toprule
\textbf{Hyper-parameter}      & \textbf{FB15K-237} & \textbf{WN18RR} \\ \midrule
Embedding dim        & 64        & 32         \\
Paths per entity     & 2         & 2    \\
Path Transformer dim       & 128  & 128         \\
Path Transformer heads     & 4   & 4         \\
Path Transformer layers    & 1   & 2    \\
Path Transformer dropout   & 0.2 & 0.1   \\
Entity aggregation   & Average       & Transformer Enc \\
Aggregation layers  &  -          & 1   \\
Loss function        & CE        & CE   \\
Learning rate        & 1e-3      & 1e-3  \\
Batch size           & 512       & 512     \\
Accumulated batches  & 8        & 8    \\
Label smoothing      & 0.1         & 0.01    \\ \midrule
Parameter count      & 0.86      & 0.05   \\
Effi = $MRR/P(M)$    & 1.13     & 17.48   \\ \bottomrule
\end{tabular}
\caption{PathE hyper-parameters for relation prediction.}
\label{tab:path-confs-rp}
\end{table}

%% file: tables/indlinkpred.tex
\begin{table*}[t]
\centering
\begin{tabular}{@{}clllllllll@{}}
\toprule
\multirow{2}{*}{\textbf{Class}}          & \multicolumn{1}{c}{\multirow{2}{*}{\textbf{Method}}} & \multicolumn{4}{c}{\textbf{FB15k-237}} & \multicolumn{4}{c}{\textbf{WN18RR}} \\ \cmidrule(l){3-10} 
                                         & \multicolumn{1}{c}{}                                 & V1       & V2      & V3      & V4      & V1      & V2      & V3     & V4     \\ \midrule
\multirow{4}{*}{Path}                    & Neural LP                                            & 0.529    & 0.589   & 0.529   & 0.559   & 0.744   & 0.689   & 0.462  & 0.671  \\
                                         & DRUM                                                 & 0.529    & 0.587   & 0.529   & 0.559   & 0.744   & 0.689   & 0.462  & 0.671  \\
                                         & RuleN                                                & 0.498    & 0.778   & 0.877   & 0.856   & 0.809   & 0.783   & 0.534  & 0.716  \\
                                         & PathE                                                & \underline{0.834}    & \underline{0.872}   & \underline{0.868}   & \underline{0.850}   & \underline{0.530}   & \underline{0.551}   & \underline{0.540}  & \underline{0.537} \\ \midrule
\multicolumn{1}{l}{\multirow{3}{*}{GNN}} & GraIL                                                & 0.642    & 0.818   & 0.828   & 0.893   & 0.825   & 0.787   & 0.584  & 0.734  \\
\multicolumn{1}{l}{}                     & NBFNet                                               & 0.834    & \textbf{0.949}   & \textbf{0.951}   & \textbf{0.960}   & \textbf{0.948}   & \textbf{0.905}   & \textbf{0.893}  & \textbf{0.890}  \\
\multicolumn{1}{l}{}                     & NP+CompGCN                                           & \textbf{0.873}    & 0.939   & 0.944   & 0.949   & 0.830   & 0.886   & 0.785  & 0.807  \\ \bottomrule
\end{tabular}
\caption{PathE \textbf{inductive} link prediction results. We report the Hits@10 with the best results in bold and PathE's results underlined. Results of other models are taken from \protect\cite{nodepiece}}
\label{tab:inductive-results}
\end{table*}

%% file: tables/inductive_properties.tex
\begin{table*}[t]
\centering
\begin{tabular}{@{}lcccccccccccc@{}}
\toprule
\multirow{2}{*}{Dataset} & \multirow{2}{*}{ } & \multirow{2}{*}{Relations} & \multicolumn{2}{c}{Train} & \multicolumn{4}{c}{Validation} & \multicolumn{3}{c}{Test} \\
 &  &  & Entity & Query & Triples & Entity & Query & Triples & Entity & Query & Triples \\ \midrule
\multirow{4}{*}{FB15k-237} & v1 & 183 & 2,000 & 4,245 & 4,245 & 1,500 & 206 & 1,993 & 1,500 & 205 & 1,993 \\
 & v2 & 203 & 3,000 & 9,739 & 9,739 & 2,000 & 469 & 4,145 & 2,000 & 478 & 4,145 \\
 & v3 & 218 & 4,000 & 17,986 & 17,986 & 3,000 & 866 & 7,406 & 3,000 & 865 & 7,406 \\
 & v4 & 222 & 5,000 & 27,203 & 27,203 & 3,500 & 1,416 & 11,714 & 3,500 & 1,424 & 11,714 \\ \midrule
\multirow{4}{*}{WN18RR} & v1 & 9 & 2,746 & 5,410 & 5,410 & 922 & 185 & 1,618 & 922 & 188 & 1,618 \\
 & v2 & 10 & 6,954 & 15,262 & 15,262 & 2,923 & 411 & 4,011 & 2,923 & 441 & 4,011 \\
 & v3 & 11 & 12,078 & 25,901 & 25,901 & 5,084 & 538 & 6,327 & 5,084 & 605 & 6,327 \\
 & v4 & 9 & 3,861 & 7,940 & 7,940 & 7,208 & 1,394 & 12,334 & 7,208 & 1,429 & 12,334 \\ \bottomrule
\end{tabular}
\caption{Dataset statistics for inductive link prediction. The term "triples" refers to the size of the input graph, while "queries" represent the triples to be predicted. In the training phase, all queries are included as triples. It is important to note that during validation and testing, a new graph, disjoint from the training graph, is provided, and queries are evaluated against this new inference graph. Consequently, the number of entities and triples remains identical for the validation and test sets as they both refer to the inference graph.}
\label{tab:inductive-stats}
\end{table*}